\documentclass[10pt,conference]{IEEEtran}
\IEEEoverridecommandlockouts
\usepackage{cite}
\usepackage{amsmath,amssymb,amsfonts}
\usepackage{algorithmic}
\usepackage{graphicx}
\usepackage{hyperref}
\usepackage{amsmath}
\usepackage{textcomp}
\usepackage{xcolor}
\usepackage{comment}
\usepackage{float}

\usepackage{algorithm} % Required for the floating algorithm environment
\usepackage{booktabs}  % Required for \toprule, \midrule, \bottomrule
\usepackage{multirow}  % Required for complex table layouts
\usepackage{url}       % Required for URL in references
\usepackage{rotating}  % For rotated column headers
\usepackage{makecell}  % For multi-line cells
\usepackage{float}     % Required for precise figure placement
\def\BibTeX{{\rm B\kern-.05em{\sc i\kern-.025em b}\kern-.08em
    T\kern-.1667em\lower.7ex\hbox{E}\kern-.125emX}}

\begin{document}

\title{Active In-Context Learning \\for Tabular Foundation Models}

\author{\IEEEauthorblockN{Wilailuck Treerath and Fabrizio Pittorino}
\IEEEauthorblockA{Dipartimento di Elettronica, Informazione e Bioingegneria,
Politecnico di Milano, Milano, Italy\\
Email: fabrizio.pittorino@polimi.it}
}

\maketitle

\begin{abstract}
Active learning (AL) reduces labeling cost by querying informative samples, but in tabular settings its cold-start gains are often limited because uncertainty estimates are unreliable when models are trained on very few labels. Tabular foundation models such as TabPFN provide calibrated probabilistic predictions via in-context learning (ICL), i.e., without task-specific weight updates, enabling an AL regime in which the labeled context - rather than parameters - is iteratively optimized. We formalize Tabular Active In-Context Learning (Tab-AICL) and instantiate it with four acquisition rules: uncertainty (TabPFN-Margin), diversity (TabPFN-Coreset), an uncertainty–diversity hybrid (TabPFN-Hybrid), and a scalable two-stage method (TabPFN-Proxy-Hybrid) that shortlists candidates using a lightweight linear proxy before TabPFN-based selection. Across 20 classification benchmarks, Tab-AICL improves cold-start sample efficiency over retrained gradient-boosting baselines (CatBoost-Margin and XGBoost-Margin), measured by normalized AULC up to 100 labeled samples. 
%We find no universally dominant acquisition rule: margin excels on clean, structured datasets; hybrid selection benefits clustered geometries; and proxy-hybrid is more robust under noisy or imbalanced pools where uncertainty sampling can fixate on outliers.
\end{abstract}


\section{Introduction}
Active learning (AL) aims to reduce annotation cost by selecting informative samples to label \cite{b1,b2}. 
In tabular classification, however, AL often provides limited benefit in the cold-start regime because acquisition functions depend on predictive uncertainty, which is unstable when models are fit on very small labeled sets. 

Tabular foundation models such as TabPFN \cite{b4, b5} provide strong probabilistic predictions via in-context learning (ICL), i.e., without task-specific gradient-based training. 
This changes the AL perspective: instead of repeatedly re-training a model after each query, we can treat the labeled set as a \emph{context} that is iteratively refined to improve predictions. 
Yet, using ICL inside an AL loop raises two practical issues. 
First, TabPFN inference scales quadratically with context length, making pool-wide scoring expensive for large unlabeled sets. 
Second, uncertainty-based querying can be brittle on noisy or imbalanced pools, where high uncertainty may correspond to outliers rather than informative boundary points.

We propose \emph{Tabular Active In-Context Learning (Tab-AICL)}, a framework that instantiates common acquisition principles within TabPFN-based ICL. 
We consider four acquisition rules: uncertainty sampling (TabPFN-Margin), diversity sampling (TabPFN-Coreset), an uncertainty--diversity hybrid (TabPFN-Hybrid), and a scalable two-stage variant (TabPFN-Proxy-Hybrid) that shortlists candidates with a lightweight logistic-regression proxy before TabPFN-based selection.

We evaluate Tab-AICL on 20 tabular classification benchmarks with 10 random seeds, focusing on cold-start performance up to 100 labeled samples. 
Across datasets, Tab-AICL improves normalized area under the learning curve (AULC) over strong retrained baselines with fixed hyperparameters (CatBoost-Margin and XGBoost-Margin) on a majority of tasks. 
Consistent with the diversity of tabular data regimes, no single acquisition rule dominates: margin sampling is strong on clean/structured datasets, hybrid selection often helps on clustered geometries, and proxy-hybrid is more robust when uncertainty sampling tends to fixate on outliers.

%%%%%%%%%%%%%%%%%%%%%%%%%%%%%%%%%%%%%%%%%%%%%%%%%%%%%%%%%%%%

\section{Background and Related Work}

\paragraph{Tabular Deep Learning and Foundation Models}
Despite major advances of deep learning in vision and language, supervised learning on tabular data is still commonly addressed with tree-based ensembles \cite{b7,b8}. 
A recurring difficulty is that tabular datasets combine heterogeneous feature types and often operate in moderate-data regimes, where generic neural architectures can be less competitive without careful inductive bias and tuning \cite{b6,b9}. 
Tabular foundation models such as TabPFN \cite{b4, b5} take a different route: a Transformer is pretrained on large collections of synthetic tabular tasks and then applied to a new dataset via in-context learning, producing probabilistic predictions without task-specific gradient-based training. 
This makes TabPFN particularly relevant for cold-start regimes, where reliable probability estimates are critical for downstream decision-making.

\paragraph{Active Learning Challenges}
Classical AL selects queries using acquisition functions derived from model predictions, most commonly uncertainty-based criteria such as least confidence, margin, or entropy \cite{b1}. 
%These strategies can be effective when predictive uncertainty tracks epistemic uncertainty, but can degrade when uncertainty is dominated by noise or outliers. 
An alternative is diversity- or geometry-driven selection, such as core-set (greedy $k$-center) methods \cite{b10}, which aim to cover the input space but may under-sample informative boundary regions when class structure is complex. 
A large body of work in deep AL addresses the uncertainty--diversity trade-off with batch-aware objectives (e.g., BatchBALD \cite{b11} and BADGE \cite{b12}), often at substantial computational cost.

\paragraph{Foundation Models and Active Learning} Recent work has begun exploring AL with foundation models in vision and language, where pretrained representations improve acquisition-function quality \cite{deepALsurvey, margatina-etal-2023-active,citovsky2021batch}. In the tabular domain, this intersection remains largely unexplored: TabPFN has been benchmarked as a predictor~\cite{b4, b5} but not systematically studied within an AL loop. Our work fills this gap by formalizing the AL-ICL coupling in the tabular foundation model context and providing a controlled empirical comparison.

Our work sits at the intersection of these lines: we study AL when the predictor is an in-context tabular foundation model. 
Rather than introducing a new Bayesian batch objective, we instantiate lightweight uncertainty, diversity, and hybrid acquisitions within TabPFN inference, and add a simple proxy-based shortlist to reduce pool-scoring cost in practice.

%%%%%%%%%%%%%%%%%%%%%%%%%%%%%%%%%%%%%%%%%%%%%%%

\section{Tabular Active In-Context Learning}
\label{sec:III}
Tab-AICL instantiates AL with an in-context tabular foundation model: the predictor is fixed, and learning proceeds by expanding the labeled context given at inference time. 
The active loop selects which labeled examples to acquire and add to the context, rather than updating model parameters.

\subsection{Problem Formulation}
Let $\mathcal{U}=\{x_1,\dots,x_M\}$ denote an unlabeled pool and $\mathcal{L}_0$ a small labeled set. 
At iteration $t$, an acquisition rule selects a batch $\mathcal{Q}_t \subset \mathcal{U}$ of size~$B$ to be labeled by an oracle. 
The labeled context is updated as
$%\begin{equation}
\mathcal{L}_{t+1} \;=\; \mathcal{L}_t \cup \{(x,y)\;:\; x\in \mathcal{Q}_t\},
$ % \end{equation}
and queried points are removed from the pool. 
Given a fixed model~$\mathcal{M}$ capable of in-context learning (TabPFN), the objective is to maximize predictive performance on a held-out test set as a function of the labeling budget $|\mathcal{L}_t|$.

\subsection{Data Preprocessing}
We apply a single preprocessing pipeline $\Phi(\cdot)$ to all methods. 
Features are treated as categorical or numerical using dataset metadata; if metadata are unavailable, integer-valued features with less than 20 unique values are treated as categorical.
Numerical features are mean-imputed (computed on the training pool) and standardized with \texttt{StandardScaler}. 
Categorical features are mode-imputed and encoded with \texttt{OrdinalEncoder}.
All preprocessing parameters are fit on the training pool only (before querying) and then applied to both $\mathcal{U}$ and the test set to avoid leakage.
The feature counts in Table~\ref{tab:datasets} refer to the dimensionality after preprocessing.

%%%%%%%%%%%%%%%%%%

\subsection{Acquisition Strategies}
We consider four acquisition rules that emphasize (i)~uncertainty, (ii)~diversity, or (iii)~their combination under computational constraints.

\subsubsection{Margin Sampling (\textbf{TabPFN-Margin})}
Margin sampling prioritizes points for which the model is least confident among its top predictions. %
%\footnote{For binary classification, margin sampling is equivalent to selecting points with posterior closest to $0.5$.}
Given a probabilistic predictor $\mathcal{M}$ conditioned on context $\mathcal{L}$, let
$p_k(x)=P(\hat{y}=k \mid x,\mathcal{L})$ denote the predicted class probabilities and let
$p_{(1)}(x)\ge p_{(2)}(x)\ge\cdots$ be these probabilities sorted in descending order.
The margin score is
$% \begin{equation}
\alpha_{\text{margin}}(x) \;=\; p_{(1)}(x) - p_{(2)}(x),
$ %\end{equation}
and we query the $B$ points with smallest $\alpha_{\text{margin}}(x)$.

\begin{algorithm}[t]
\caption{Tab-AICL Margin Sampling (\textbf{TabPFN-Margin})}
\label{alg:margin}
\begin{algorithmic}[1]
\REQUIRE Unlabeled pool $\mathcal{U}$, labeled context $\mathcal{L}$, batch size $B$, model $\mathcal{M}$, budget $N_{\max}$
\WHILE{$|\mathcal{L}| < N_{\max}$ \AND $|\mathcal{U}|>0$}
    \STATE $P \leftarrow \mathcal{M}.\texttt{predict\_proba}(\mathcal{U};\,\mathcal{L})$
    \STATE \textbf{for} each $x_i\in\mathcal{U}$ compute $\alpha_i \leftarrow p_{(1)}(x_i)-p_{(2)}(x_i)$ from $P_i$
    \STATE $\mathcal{Q} \leftarrow \texttt{arg\,sort}(\alpha)[:\min(B,|\mathcal{U}|)]$ \hfill \textit{// smallest margins}
    \STATE obtain labels for $\mathcal{Q}$ and update $\mathcal{L}\leftarrow \mathcal{L}\cup \texttt{Oracle}(\mathcal{Q})$
    \STATE update $\mathcal{U}\leftarrow \mathcal{U}\setminus \mathcal{Q}$
\ENDWHILE
\end{algorithmic}
\end{algorithm}

\subsubsection{Hybrid Sampling (\textbf{TabPFN-Hybrid})}
Hybrid sampling combines an uncertainty filter with a diversity step. The intuition is to (i) restrict attention to a
candidate set that the current predictor finds ambiguous, and then (ii) select a geometrically diverse batch from that
candidate set. Concretely, we compute predictive entropy
%\begin{equation}
%\begin{aligned}
$H(x)=-\sum_{k=1}^{K} p_k(x)\,\log p_k(x)$, with $p_k(x)=P(\hat{y}=k \mid x,\mathcal{L})$,
%\label{eq:entropy}
%\end{aligned}
%\end{equation}
retain the $N_{\text{cand}}$ most entropic points, and run $k$-means ($k=B$) on their preprocessed feature vectors.
We then query the point nearest to each centroid.

\begin{algorithm}[h!]
\caption{Tab-AICL Hybrid Sampling (\textbf{TabPFN-Hybrid})}
\label{alg:hybrid}
\begin{algorithmic}[1]
\REQUIRE Unlabeled pool $\mathcal{U}$, labeled context $\mathcal{L}$, batch size $B$, model $\mathcal{M}$, budget $N_{\max}$
\WHILE{$|\mathcal{L}| < N_{\max}$ \AND $|\mathcal{U}|>0$}
    \STATE $P \leftarrow \mathcal{M}.\texttt{predict\_proba}(\mathcal{U};\,\mathcal{L})$
    \STATE \textbf{for} each $x_i\in\mathcal{U}$ compute $H_i \leftarrow -\sum_{k=1}^K p_{ik}\log(p_{ik}+\varepsilon)$
    \STATE $N_{\text{cand}} \leftarrow \min\big(|\mathcal{U}|,\;\max(2B,\lfloor |\mathcal{U}|/2 \rfloor)\big)$
    \STATE $\mathcal{S} \leftarrow \texttt{topk}(H, N_{\text{cand}})$ \hfill \textit{// highest entropy indices}
    \STATE $X \leftarrow \{x_i : i\in\mathcal{S}\}$ \hfill \textit{// in preprocessed feature space}
    \STATE fit $k$-means with $k=B$ on $X$ (if $|X|<B$, $k\leftarrow |X|$)
    \STATE $\mathcal{Q} \leftarrow$ nearest point to each centroid (unique selection; break ties arbitrarily)
    \STATE obtain labels for $\mathcal{Q}$ and update $\mathcal{L}\leftarrow \mathcal{L}\cup \texttt{Oracle}(\mathcal{Q})$
    \STATE update $\mathcal{U}\leftarrow \mathcal{U}\setminus \mathcal{Q}$
\ENDWHILE
\end{algorithmic}
\end{algorithm}

\subsubsection{Proxy-Hybrid Sampling (\textbf{TabPFN-Proxy-Hybrid})}
The Proxy-Hybrid strategy reduces acquisition cost by screening the pool with a lightweight proxy and invoking TabPFN only
on a shortlisted subset. At each iteration, we (i) fit a linear classifier $\mathcal{P}$ on the current labeled context
$\mathcal{L}$, (ii) rank all $x\in\mathcal{U}$ by the proxy predictive entropy, and (iii) retain the top
$N_{\text{proxy}}$ candidates, where $N_{\text{proxy}}$ is clamped to bound runtime:
\begin{equation}
N_{\text{proxy}} \;=\; \min\!\Big(|\mathcal{U}|,\; \max\big(N_{\min},\, \min(N_{\max},\, \alpha|\mathcal{U}|)\big)\Big),
\label{eq:nproxy}
\end{equation}
with $N_{\min}=200$, $N_{\max}=2000$, and $\alpha=0.05$ in our experiments. We then evaluate TabPFN uncertainty only on
this shortlist and apply a diversity step to form the batch.

We use entropy for both stages. For a distribution $p(\cdot\mid x)$, define
$H(x)=-\sum_{k=1}^K p_k(x)\log(p_k(x)+\varepsilon)$ with $\varepsilon>0$.
Let $\mathcal{S}_{\text{short}}=\texttt{topk}(H_{\mathcal{P}}, N_{\text{proxy}})$ be the proxy shortlist. We compute
TabPFN entropy on $\mathcal{S}_{\text{short}}$, retain the $N_{\text{div}}=\min(3B,|\mathcal{S}_{\text{short}}|)$
highest-entropy points, and run $k$-means with $k=B$ on their preprocessed features. The queried batch $\mathcal{Q}$
contains the nearest point to each centroid (with unique selection).

\paragraph*{Proxy configuration and cost}
In TabPFN-Proxy-Hybrid, the proxy $\mathcal{P}$ is a logistic regression classifier with balanced class weights.
At each acquisition step, $\mathcal{P}$ scores all $x\in\mathcal{U}$ and we retain a shortlist
$\mathcal{S}_{\text{short}}$ containing the top $\alpha$ fraction of points by proxy predictive entropy
($\alpha=0.05$ in our experiments), subject to the clamping in~\eqref{eq:nproxy}.
TabPFN uncertainty and the subsequent diversity step are computed only on $\mathcal{S}_{\text{short}}$,
reducing the number of expensive TabPFN pool evaluations by approximately a factor $1/\alpha$.
\begin{algorithm}[t!]
\caption{Tab-AICL Proxy-Hybrid Sampling (\textbf{TabPFN-Proxy-Hybrid})}
\label{alg:proxy}
\begin{algorithmic}[1]
\REQUIRE Unlabeled pool $\mathcal{U}$, labeled context $\mathcal{L}$, batch size $B$, model $\mathcal{M}$, proxy $\mathcal{P}$,
budget $N_{\max}$, filter ratio $\alpha$, clamps $(N_{\min},N_{\max}^{proxy})$
\WHILE{$|\mathcal{L}| < N_{\max}$ \AND $|\mathcal{U}|>0$}
    \STATE fit proxy $\mathcal{P}$ on $\mathcal{L}$
    \STATE compute proxy entropy $H_{\mathcal{P}}(x)$ for all $x\in\mathcal{U}$
    \STATE $N_{\text{proxy}} \leftarrow \min\!\Big(|\mathcal{U}|,\max\big(N_{\min},\min(N_{\max}^{proxy},\lfloor \alpha|\mathcal{U}|\rfloor)\big)\Big)$
    \STATE $\mathcal{S}_{\text{short}} \leftarrow \texttt{topk}(H_{\mathcal{P}}, N_{\text{proxy}})$
    \STATE compute TabPFN entropy $H_{\mathcal{M}}(x)$ for $x\in\mathcal{S}_{\text{short}}$ using context $\mathcal{L}$
    \STATE $N_{\text{div}} \leftarrow \min(3B,|\mathcal{S}_{\text{short}}|)$
    \STATE $\mathcal{S}_{\text{final}} \leftarrow \texttt{topk}(H_{\mathcal{M}}, N_{\text{div}})$
    \STATE run $k$-means with $k=\min(B,|\mathcal{S}_{\text{final}}|)$ on $\{x_i: i\in\mathcal{S}_{\text{final}}\}$
    \STATE $\mathcal{Q} \leftarrow$ nearest point to each centroid (unique selection; break ties arbitrarily)
    \STATE update $\mathcal{L}\leftarrow \mathcal{L}\cup \texttt{Oracle}(\mathcal{Q})$, \;\; $\mathcal{U}\leftarrow \mathcal{U}\setminus \mathcal{Q}$
\ENDWHILE
\end{algorithmic}
\end{algorithm}

\subsubsection{Coreset Sampling (Diversity)}
We adopt a greedy $k$-center (coreset) rule that promotes coverage of the input space by selecting points that are far from the current labeled set in the standardized feature space.
Let $z(\cdot)$ denote the preprocessing map and $d(\cdot,\cdot)$ the Euclidean distance.
At each selection, we pick
%\begin{equation}
$x^* = \arg\max_{u\in\mathcal{U}} \; \min_{l\in\mathcal{L}} \, \lVert z(u)-z(l)\rVert_2$,
%\label{eq:coreset}
%\end{equation}
and repeat this procedure $B$ times to form the query batch.
To avoid storing the full $|\mathcal{U}|\times|\mathcal{L}|$ distance matrix, we maintain a vector of current nearest-neighbor distances
$D_{\min}(u)=\min_{l\in\mathcal{L}}\lVert z(u)-z(l)\rVert_2$ for all $u\in\mathcal{U}$ and update it incrementally after each newly selected point.
This yields $\mathcal{O}(|\mathcal{U}|)$ memory, with per-step time dominated by computing distances from the new center to all remaining unlabeled points.
\begin{algorithm}[htbp]
\caption{Tab-AICL Coreset (Greedy $k$-Center) (\textbf{TabPFN-Coreset})}
\label{alg:coreset}
\begin{algorithmic}[1]
\REQUIRE Unlabeled pool $\mathcal{U}$, labeled context $\mathcal{L}$, batch size $B$, budget $N_{\max}$
\ENSURE Updated labeled context $\mathcal{L}$
\WHILE{$|\mathcal{L}| < N_{\max}$}
    \STATE \textit{// Maintain nearest-labeled distance for each } $u\in\mathcal{U}$
    \STATE $D_{\min}(u) \leftarrow \min_{l\in\mathcal{L}} \|z(u)-z(l)\|_2 \quad \forall u\in\mathcal{U}$
    \STATE $\mathcal{Q} \leftarrow \emptyset$
    \FOR{$j \leftarrow 1$ \TO $B$}
        \STATE $q \leftarrow \arg\max_{u\in\mathcal{U}} D_{\min}(u)$
        \STATE $\mathcal{Q} \leftarrow \mathcal{Q} \cup \{q\}$
        \STATE \textit{// Update distances after adding new center $q$}
        \FORALL{$u \in \mathcal{U}\setminus\{q\}$}
            \STATE $D_{\min}(u) \leftarrow \min\!\left(D_{\min}(u), \|z(u)-z(q)\|_2\right)$
        \ENDFOR
        \STATE $\mathcal{U} \leftarrow \mathcal{U} \setminus \{q\}$
    \ENDFOR
    \STATE $\mathcal{L} \leftarrow \mathcal{L} \cup \text{Oracle}(\mathcal{Q})$
\ENDWHILE
\end{algorithmic}
\end{algorithm}

%%%%%%%%%%%%%%%%%%%%%%%%%%%%%%%%%%%%%%%%%%%%%%%%%%%%%%%
\section{Experimental Setup}

\subsection{Active Learning Protocol}
We simulate pool-based batch AL under a fixed labeling budget. For each dataset, we create a stratified 70/30 split into an unlabeled training pool $\mathcal{U}$ and a held-out test set, which is never used for acquisition or preprocessing decisions. The initial labeled context $\mathcal{L}_0$ contains one randomly sampled instance per class. At each round, an acquisition rule selects a batch $\mathcal{Q}_t \subset \mathcal{U}$ of size $B$ without replacement, an oracle reveals the labels, and we update $\mathcal{L}_{t+1}=\mathcal{L}_t \cup \{(x,y):x\in\mathcal{Q}_t\}$ and $\mathcal{U}\leftarrow \mathcal{U}\setminus \mathcal{Q}_t$. The loop terminates when $|\mathcal{L}_t| = N_{\max}=100$. 
For very small datasets, we additionally stop once 50\% of the original unlabeled pool is queried.
%, even if the total count is less than~$100$.
%to avoid degenerate regimes where most points are labeled.

\subsection{Experimental Design and Evaluation}
We repeat the full pipeline for 10 random seeds, resampling both the split and the initialization, and report mean $\pm$ std. We evaluate batch sizes $B \in \{5,10,15,20\}$ to probe the trade-off between update frequency and within-batch diversity. To compare the cold-start efficiency of different methodologies, we report the normalized area under the learning curve (AULC) up to $N_{\max}$ as the primary metric. Let $y_t$ denote the test-set Cohen~$\kappa$ after round~$t$, when $|\mathcal{L}_t|=n_t$. We compute
\begin{equation}
\mathrm{AULC}_{\mathrm{norm}}
= \frac{1}{N_{\max}-n_0}
\sum_{t=1}^{T}
\frac{y_t+y_{t-1}}{2}\,(n_t-n_{t-1}),
\end{equation}
which corresponds to the average $\kappa$ over the cold-start acquisition trajectory. We also report final $\kappa$ at $N_{\max}$ and ROC AUC; for multi-class problems, ROC AUC is computed one-vs-rest with macro-averaging. 
% We compute the ROC AUC using the One-vs-Rest (OvR) strategy for multi-class datasets, averaging the scores using a macro-average to treat all classes equally.
For significance testing, we use paired Wilcoxon signed-rank tests on per-seed $\mathrm{AULC}_{\mathrm{norm}}$ scores. To control multiplicity across datasets, we apply Benjamini-Hochberg FDR correction and consider differences significant with the BH-adjusted $p$-value $p_{\mathrm{adj}}<0.05$.
%To evaluate performance differences without assuming data normality, we employ the non-parametric \textbf{Wilcoxon Signed-Rank Test} ($N_{seeds}=10$) on paired AULC scores. Given the exploratory nature of benchmarking across 20 diverse datasets, we control the False Discovery Rate (FDR) using the \textbf{Benjamini-Hochberg procedure}. This approach provides a rigorous standard for significance ($p_{adj} < 0.05$).

\subsection{Baselines and Reproducibility}
All methods are evaluated within the same AL protocol (identical splits, initialization $\mathcal{L}_0$, budget, and query-without-replacement). For TabPFN-based methods, we use TabPFN (\texttt{tabpfn} Python package v6.2.0) with $n_{\mathrm{estimators}}=32$ throughout. We report four Tab-AICL acquisition rules (TabPFN-Margin, TabPFN-Coreset, TabPFN-Hybrid, TabPFN-Proxy-Hybrid) and a passive TabPFN-Random baseline that uses the same TabPFN configuration but selects $\mathcal{Q}_t$ uniformly at random. The oracle is simulated: ground-truth labels from the dataset are revealed upon query.

As retrained baselines, we include CatBoost and XGBoost with margin-based querying, where the model is refit from scratch at each round on the current labeled set $\mathcal{L}_t$. 
To avoid per-dataset tuning, which would be impractical under a realistic cold-start protocol, we fix hyperparameters for stability in low-data regimes. $n_{\mathrm{estimators}}=2000$, $\mathrm{learning\_rate}=0.05$, $\mathrm{max\_depth}=6$, and $\mathrm{early\_stopping\_rounds}=50$. 
We use \texttt{scikit-learn} for preprocessing and logistic regression, and the official \texttt{catboost} and \texttt{xgboost} packages. 
Finally, we include a semi-supervised Label Spreading baseline \cite{b15} with a kNN kernel ($k=7$) and clamping factor $\alpha=0.2$, paired with random acquisition to isolate the effect of the learner from the query rule.
%Experiments are run on a workstation with an Intel Core i9-14900K CPU (24 Cores, 32 Logical Processors) and 64\,GB RAM, using PyTorch 2.9.1.

\subsection{Datasets}
We benchmark 20 classification datasets from OpenML~\cite{OpenML} and UCI~\cite{UCI} (Table~\ref{tab:datasets}), spanning heterogeneous domains and mixed feature types. To keep runtime comparable across benchmarks and reflect practical labeling budgets, datasets with more than 10{,}000 instances are uniformly subsampled to 10{,}000 prior to splitting. All datasets are processed with the standardized pipeline described in Section~\ref{sec:III}, and the reported feature counts correspond to the post-preprocessing dimensionality.

\begin{table}[htbp]
\centering
\caption{Benchmark datasets (post-preprocessing).}
\label{tab:datasets}
\resizebox{\columnwidth}{!}{%
\begin{tabular}{lccccc}
\toprule
Dataset & ID & Instances & Features & Classes & Domain \\
\midrule
Iris & 61 & 150 & 4 & 3 & Botany \\
Glass & 41 & 214 & 9 & 6 & Forensic \\
Ionosphere & 59 & 351 & 34 & 2 & Physics \\
Balance-scale & 997 & 625 & 4 & 2 & Psychology \\
Vehicle & 994 & 846 & 18 & 2 & Image Rec. \\
Page-blocks & 1021 & 5{,}473 & 10 & 2 & Document \\
Parkinsons & 1488 & 195 & 22 & 2 & Medicine \\
Seeds & 1499 & 210 & 7 & 3 & Agriculture \\
Bank-Marketing & 45065 & 45{,}211 (sub: 10{,}000) & 16 & 2 & Finance \\
Adult & 45068 & 48{,}842 (sub: 10{,}000) & 14 & 2 & Social Sci. \\
CoverType & 1596 & 581{,}012 (sub: 10{,}000) & 54 & 7 & Forestry \\
KC1 & 1066 & 145 & 94 & 2 & Software \\
JM1 & 46979 & 10{,}885 (sub: 10{,}000) & 21 & 2 & Software \\
Blood-Transfusion & 46913 & 748 & 4 & 2 & Medicine \\
Diabetes & 46921 & 768 & 8 & 2 & Medicine \\
Tic-Tac-Toe & 137 & 39{,}366 (sub: 10{,}000) & 9 & 2 & Game Theory \\
Credit-g & 46918 & 1{,}000 & 20 & 2 & Finance \\
Steel-Plates & 46959 & 1{,}941 & 26 & 7 & Industry \\
Phoneme & 44127 & 3{,}172 & 5 & 2 & Speech \\
Ilpd & 41943 & 583 & 10 & 2 & Medicine \\
\bottomrule
\end{tabular}%
}
\vspace{0.1cm}
\parbox{\columnwidth}{\footnotesize Datasets with more than 10{,}000 instances are subsampled uniformly to 10{,}000 prior to splitting.}
\end{table}

%%%%%%%%%%%%%%%%%%%%%%%%%%%%%%%%%%%%%%%%%%%%%%%%%%%%%
\section{Results and Discussion}

\subsection{Cold-start efficiency vs.\ baselines}
Table~\ref{tab:performance} reports normalized AULC up to 100 labels. In this regime, TabPFN provides a strong cold-start baseline and, when combined with active acquisition, attains the best AULC on 15/20 datasets. Overall, TabPFN-based methods outperform the retrained GBDT baselines on 18/20 datasets, suggesting that ICL inference provides a favorable inductive bias in the low-label setting. Nevertheless, active querying is not uniformly beneficial: on \texttt{Glass}, \texttt{Bank-Marketing}, and \texttt{Tic-Tac-Toe}, TabPFN-Random matches or exceeds the active rules. 
%, consistent with settings where uncertainty-driven selection can over-focus on atypical or noisy points
Conversely, CatBoost-Margin remains competitive on \texttt{KC1} and \texttt{Ilpd}, indicating that tree-based biases can still be effective for some tabular datasets even under tight label budgets.
For completeness, we report the corresponding final-step performance at the 100-label budget (Cohen's $\kappa$ and ROC AUC) in Appendix~\ref{app:final_metrics}.

%\subsection{Statistical significance verification}
We test whether TabPFN-Hybrid improves over the strongest retrained GBDT baseline (CatBoost-Margin) using a paired Wilcoxon signed-rank test across seeds ($N_{\text{seeds}}=10$) on normalized AULC scores. Table~\ref{tab:significance} reports per-dataset differences $\Delta\text{AULC}$ and the corresponding $p$-values after Benjamini--Hochberg correction over the 20 datasets. After correction, TabPFN-Hybrid is significantly better on 8/20 datasets ($p_{\mathrm{adj}}<0.05$). In the opposite direction, CatBoost-Margin does not achieve a statistically significant advantage over TabPFN-Hybrid on any dataset under the same protocol.

%\subsection{When do acquisition rules help?}
Across datasets, the benefit of active querying is not uniform and no single acquisition rule is consistently best. TabPFN-Margin is frequently competitive, while adding a diversity step (TabPFN-Hybrid) can improve performance on some benchmarks. TabPFN-Proxy-Hybrid is often among the more stable options, but it does not dominate. Finally, on several datasets TabPFN-Random matches or exceeds the active rules in the 0--100 label regime, indicating that acquisition can be dataset-sensitive under this budget and protocol.

\begin{table*}[t]
\caption{AULC Performance Summary (Mean $\pm$ std of AULC up to $N=100$, Batch Size 10)}
\label{tab:performance}
\centering
\footnotesize
\resizebox{\linewidth}{!}{%
\begin{tabular}{@{}lcccccccc@{}}
\toprule
\multirow{2}{*}{\textbf{Dataset}} & \multicolumn{5}{c}{\textbf{Tab-AICL Strategies}} & \multicolumn{3}{c}{\textbf{Baselines}} \\
\cmidrule(lr){2-5} \cmidrule(lr){6-9}
 & \textbf{TabPFN-Coreset} & \textbf{TabPFN-Hybrid} & \textbf{TabPFN-Proxy-Hybrid} & \textbf{TabPFN-Margin} & \textbf{TabPFN-Random} & \textbf{CatBoost-Margin} & \textbf{XGBoost-Margin} & \textbf{LabelSpreading-Random} \\
\midrule
Iris & \textbf{0.934 $\pm$ 0.023} & 0.929 $\pm$ 0.029 & 0.931 $\pm$ 0.019 & 0.917 $\pm$ 0.027 & 0.902 $\pm$ 0.031 & 0.880 $\pm$ 0.060 & 0.743 $\pm$ 0.074 & 0.831 $\pm$ 0.058 \\
Glass & 0.397 $\pm$ 0.055 & 0.472 $\pm$ 0.041 & 0.441 $\pm$ 0.061 & 0.481 $\pm$ 0.056 & \textbf{0.489 $\pm$ 0.077} & 0.456 $\pm$ 0.055 & 0.400 $\pm$ 0.033 & 0.382 $\pm$ 0.061 \\
Ionosphere & 0.617 $\pm$ 0.132 & \textbf{0.810 $\pm$ 0.042} & 0.800 $\pm$ 0.032 & 0.788 $\pm$ 0.038 & 0.751 $\pm$ 0.052 & 0.641 $\pm$ 0.051 & 0.640 $\pm$ 0.049 & 0.440 $\pm$ 0.154 \\
Balance-scale & 0.758 $\pm$ 0.032 & 0.808 $\pm$ 0.024 & \textbf{0.811 $\pm$ 0.040} & 0.808 $\pm$ 0.044 & 0.721 $\pm$ 0.055 & 0.609 $\pm$ 0.042 & 0.578 $\pm$ 0.058 & 0.665 $\pm$ 0.049 \\
Vehicle & 0.742 $\pm$ 0.067 & \textbf{0.833 $\pm$ 0.022} & 0.811 $\pm$ 0.054 & 0.794 $\pm$ 0.040 & 0.747 $\pm$ 0.052 & 0.665 $\pm$ 0.057 & 0.640 $\pm$ 0.062 & 0.685 $\pm$ 0.056 \\
Page-blocks & 0.602 $\pm$ 0.064 & 0.687 $\pm$ 0.044 & 0.657 $\pm$ 0.043 & \textbf{0.704 $\pm$ 0.023} & 0.624 $\pm$ 0.042 & 0.592 $\pm$ 0.038 & 0.349 $\pm$ 0.166 & 0.430 $\pm$ 0.044 \\
Parkinsons & 0.575 $\pm$ 0.126 & \textbf{0.629 $\pm$ 0.125} & 0.618 $\pm$ 0.128 & 0.580 $\pm$ 0.160 & 0.580 $\pm$ 0.125 & 0.574 $\pm$ 0.135 & 0.489 $\pm$ 0.127 & 0.520 $\pm$ 0.089 \\
Seeds & 0.891 $\pm$ 0.030 & \textbf{0.907 $\pm$ 0.014} & 0.903 $\pm$ 0.022 & 0.900 $\pm$ 0.026 & 0.893 $\pm$ 0.037 & 0.839 $\pm$ 0.047 & 0.742 $\pm$ 0.030 & 0.841 $\pm$ 0.041 \\
Bank-Marketing & 0.211 $\pm$ 0.074 & 0.175 $\pm$ 0.077 & 0.217 $\pm$ 0.100 & 0.158 $\pm$ 0.043 & \textbf{0.258 $\pm$ 0.075} & 0.148 $\pm$ 0.058 & 0.150 $\pm$ 0.060 & 0.091 $\pm$ 0.025 \\
Adult & 0.298 $\pm$ 0.077 & 0.292 $\pm$ 0.059 & \textbf{0.358 $\pm$ 0.067} & 0.317 $\pm$ 0.079 & 0.300 $\pm$ 0.099 & 0.307 $\pm$ 0.060 & 0.236 $\pm$ 0.065 & 0.131 $\pm$ 0.038 \\
CoverType & 0.349 $\pm$ 0.042 & \textbf{0.383 $\pm$ 0.029} & 0.315 $\pm$ 0.046 & 0.378 $\pm$ 0.049 & 0.378 $\pm$ 0.025 & 0.326 $\pm$ 0.046 & 0.302 $\pm$ 0.035 & 0.188 $\pm$ 0.017 \\
KC1 & 0.356 $\pm$ 0.100 & 0.391 $\pm$ 0.084 & 0.412 $\pm$ 0.103 & 0.378 $\pm$ 0.121 & 0.377 $\pm$ 0.128 & \textbf{0.417 $\pm$ 0.130} & 0.391 $\pm$ 0.090 & 0.281 $\pm$ 0.093 \\
JM1 & 0.105 $\pm$ 0.059 & \textbf{0.137 $\pm$ 0.028} & 0.106 $\pm$ 0.056 & 0.069 $\pm$ 0.071 & 0.090 $\pm$ 0.061 & 0.080 $\pm$ 0.053 & 0.049 $\pm$ 0.037 & 0.088 $\pm$ 0.046 \\
Blood-Transfusion & 0.143 $\pm$ 0.086 & 0.126 $\pm$ 0.059 & 0.126 $\pm$ 0.093 & \textbf{0.151 $\pm$ 0.101} & 0.109 $\pm$ 0.085 & 0.148 $\pm$ 0.072 & 0.091 $\pm$ 0.068 & 0.140 $\pm$ 0.063 \\
Diabetes & 0.340 $\pm$ 0.046 & 0.392 $\pm$ 0.051 & \textbf{0.401 $\pm$ 0.069} & 0.362 $\pm$ 0.053 & 0.379 $\pm$ 0.050 & 0.344 $\pm$ 0.062 & 0.288 $\pm$ 0.057 & 0.192 $\pm$ 0.077 \\
Tic-Tac-Toe & 0.071 $\pm$ 0.042 & 0.145 $\pm$ 0.053 & 0.175 $\pm$ 0.075 & 0.142 $\pm$ 0.099 & \textbf{0.189 $\pm$ 0.041} & 0.122 $\pm$ 0.068 & 0.128 $\pm$ 0.079 & 0.076 $\pm$ 0.028 \\
Credit-g & 0.163 $\pm$ 0.056 & \textbf{0.180 $\pm$ 0.070} & 0.164 $\pm$ 0.058 & 0.146 $\pm$ 0.065 & 0.116 $\pm$ 0.060 & 0.159 $\pm$ 0.082 & 0.075 $\pm$ 0.054 & 0.105 $\pm$ 0.049 \\
Steel-Plates & 0.323 $\pm$ 0.039 & \textbf{0.484 $\pm$ 0.018} & 0.350 $\pm$ 0.079 & 0.450 $\pm$ 0.037 & 0.435 $\pm$ 0.036 & 0.419 $\pm$ 0.042 & 0.342 $\pm$ 0.032 & 0.367 $\pm$ 0.031 \\
Phoneme & 0.417 $\pm$ 0.053 & \textbf{0.476 $\pm$ 0.053} & 0.440 $\pm$ 0.122 & 0.444 $\pm$ 0.083 & 0.475 $\pm$ 0.055 & 0.416 $\pm$ 0.047 & 0.372 $\pm$ 0.063 & 0.417 $\pm$ 0.051 \\
Ilpd & 0.065 $\pm$ 0.072 & 0.089 $\pm$ 0.057 & 0.109 $\pm$ 0.087 & 0.115 $\pm$ 0.069 & 0.111 $\pm$ 0.068 & \textbf{0.147 $\pm$ 0.053} & 0.112 $\pm$ 0.048 & 0.100 $\pm$ 0.060 \\
\bottomrule
\end{tabular}%
}
\vspace{0.2cm}
\parbox{\linewidth}{
\footnotesize \textbf{Note:} Bold indicates the strategy with the highest mean AULC per dataset.
All values represent Mean $\pm$ Standard Deviation across 10 random seeds.
}
\end{table*}

\begin{table}[htbp]
\caption{Statistical Significance: TabPFN-Hybrid vs. CatBoost-Margin (Wilcoxon Signed-Rank Test on AULC)}
\label{tab:significance}
\centering
\footnotesize
\resizebox{\columnwidth}{!}{%
\begin{tabular}{@{}lcccc@{}}
\toprule
\textbf{Dataset} & \textbf{$\Delta \text{AULC}$} & \textbf{$p$-Value (Raw)} & \textbf{$p$-Value (Adj)} & \textbf{Result} \\
\midrule
Iris & +0.050 & 0.037 & 0.080 & No significant difference \\
Glass & +0.016 & 0.375 & 0.526 & No significant difference \\
Ionosphere & +0.169 & 0.002 & \textbf{0.007} & \textbf{Significantly higher} \\
Balance-scale & +0.199 & 0.002 & \textbf{0.007} & \textbf{Significantly higher} \\
Vehicle & +0.168 & 0.002 & \textbf{0.007} & \textbf{Significantly higher} \\
Page-blocks & +0.095 & 0.002 & \textbf{0.007} & \textbf{Significantly higher} \\
Parkinsons & +0.054 & 0.375 & 0.526 & No significant difference \\
Seeds & +0.068 & 0.004 & \textbf{0.012} & \textbf{Significantly higher} \\
Bank-Marketing & +0.028 & 0.557 & 0.683 & No significant difference \\
Adult & -0.015 & 0.695 & 0.789 & No significant difference \\
CoverType & +0.057 & 0.002 & \textbf{0.007} & \textbf{Significantly higher} \\
KC1 & -0.027 & 0.492 & 0.635 & No significant difference \\
JM1 & +0.057 & 0.004 & \textbf{0.012} & \textbf{Significantly higher} \\
Blood-Transfusion & -0.023 & 0.375 & 0.526 & No significant difference \\
Diabetes & +0.048 & 0.037 & 0.080 & No significant difference \\
Tic-Tac-Toe & +0.023 & 0.557 & 0.683 & No significant difference \\
Credit-g & +0.020 & 0.557 & 0.683 & No significant difference \\
Steel-Plates & +0.064 & 0.004 & \textbf{0.012} & \textbf{Significantly higher} \\
Phoneme & +0.060 & 0.084 & 0.161 & No significant difference \\
Ilpd & -0.058 & 0.131 & 0.229 & No significant difference \\
\bottomrule
\end{tabular}%
}
\vspace{0.1cm}
\parbox{\columnwidth}
{\footnotesize \textbf{Note:} 
$\Delta \text{AULC}$ is the mean difference in \textbf{AULC}.
The \textbf{Result} column is determined by the Benjamini-Hochberg corrected $p$-value ($p_{adj}<0.05$).
$p_{adj}$ indicates Benjamini-Hochberg (FDR) corrected values .}
\end{table}

\subsection{Visual analysis of learning trajectories}
\begin{figure}[t]
\centering
\includegraphics[width=\columnwidth]{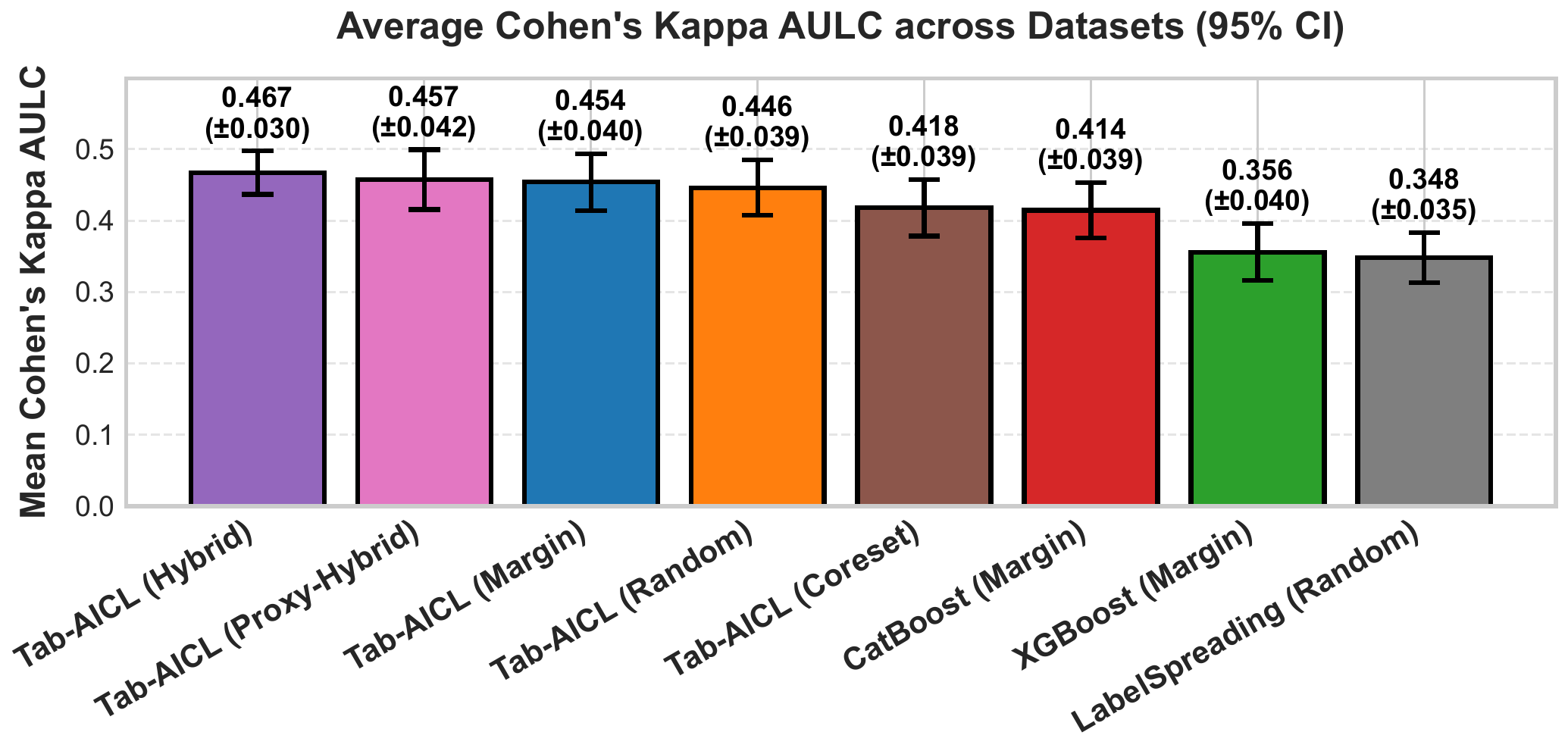}
\caption{Comprehensive AULC Performance: Area Under Learning Curve scores across 20 datasets, grouped by acquisition strategy.}
\label{fig:aulc_comprehensive}
\end{figure}

\begin{figure}[t]
\centering
\includegraphics[width=\columnwidth]{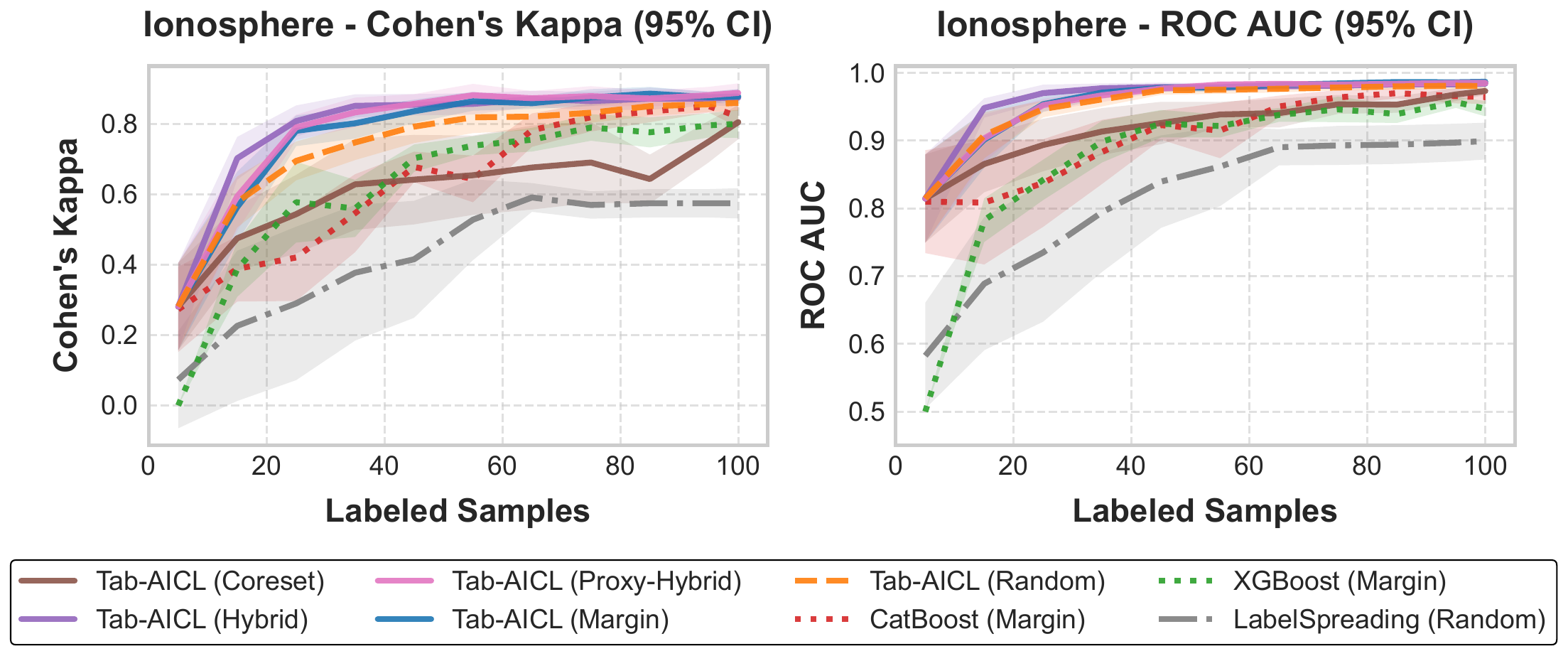}
\caption{Learning Curve for Ionosphere: Tab-AICL strategies achieve rapid convergence, with \textbf{TabPFN-Proxy-Hybrid} offering an effective balance of speed and stability.
The shaded regions represent 95\% confidence intervals across 10 random seeds.}
\label{fig:ionosphere}
\end{figure}

\begin{figure}[t]
\centering
\includegraphics[width=\columnwidth]{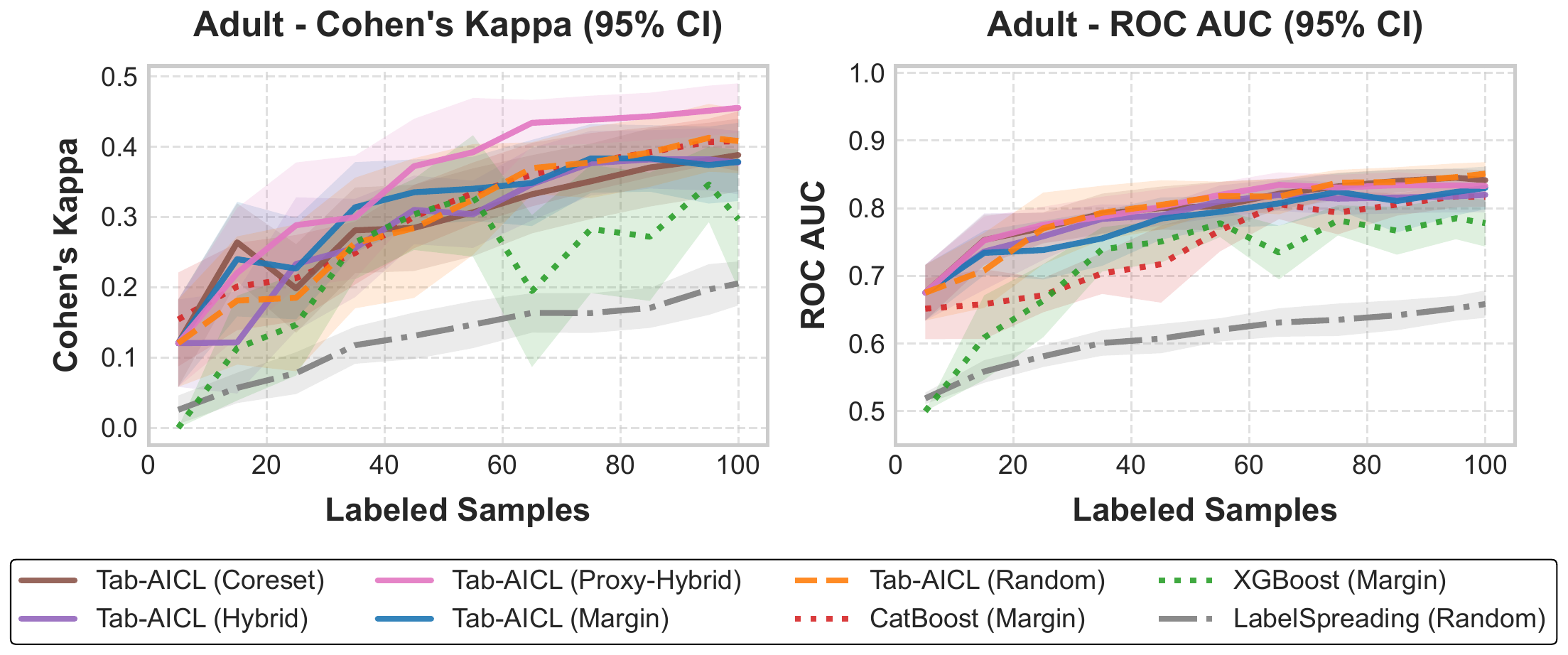}
\caption{Learning Curve for Adult: The \textbf{TabPFN-Proxy-Hybrid} strategy (pink) demonstrates stable performance and consistent gains.
Note the narrower confidence intervals compared to other strategies.}
\label{fig:adult}
\end{figure}

\begin{figure}[t]
\centering
\includegraphics[width=\columnwidth]{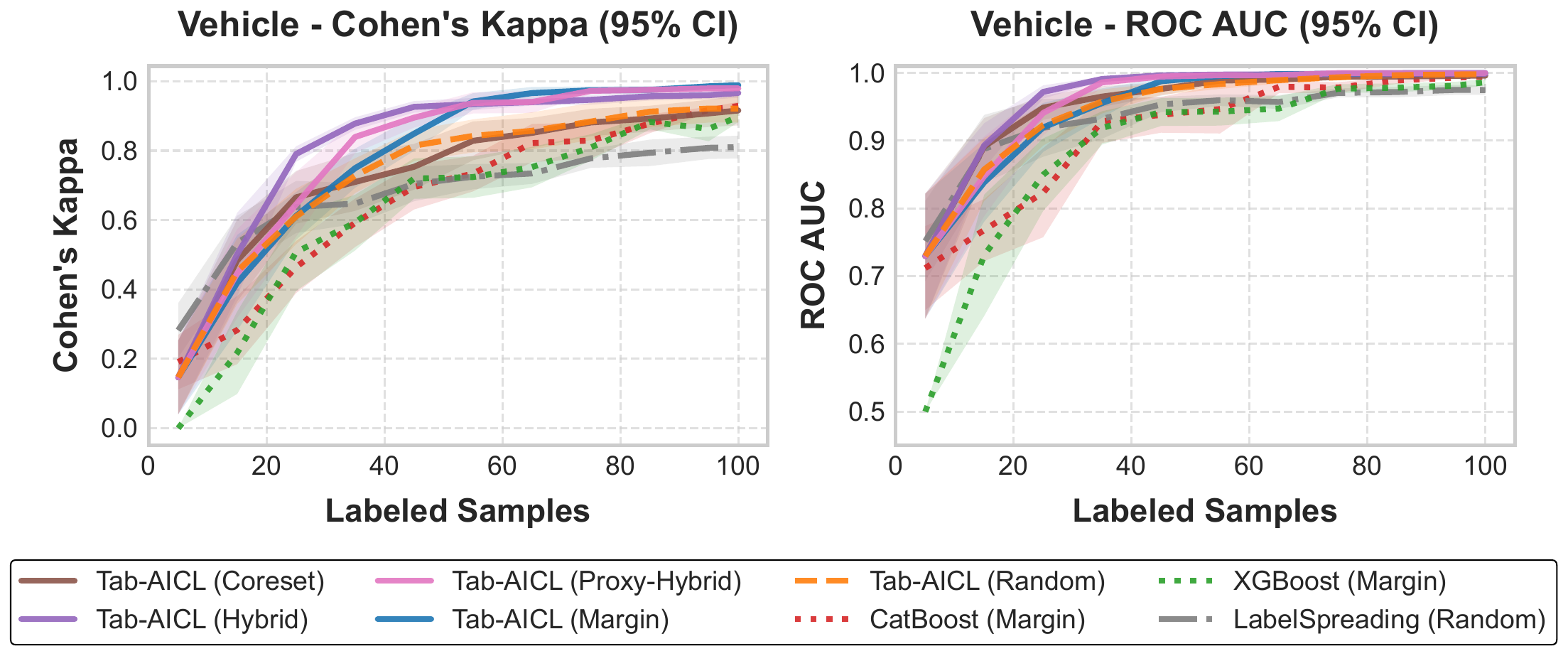}
\caption{Learning Curve for Vehicle: The \textbf{TabPFN-Hybrid} strategy (purple) achieves higher performance than pure uncertainty sampling and baselines.}
\label{fig:vehicle}
\end{figure}

\begin{figure}[t]
\centering
\includegraphics[width=\columnwidth]{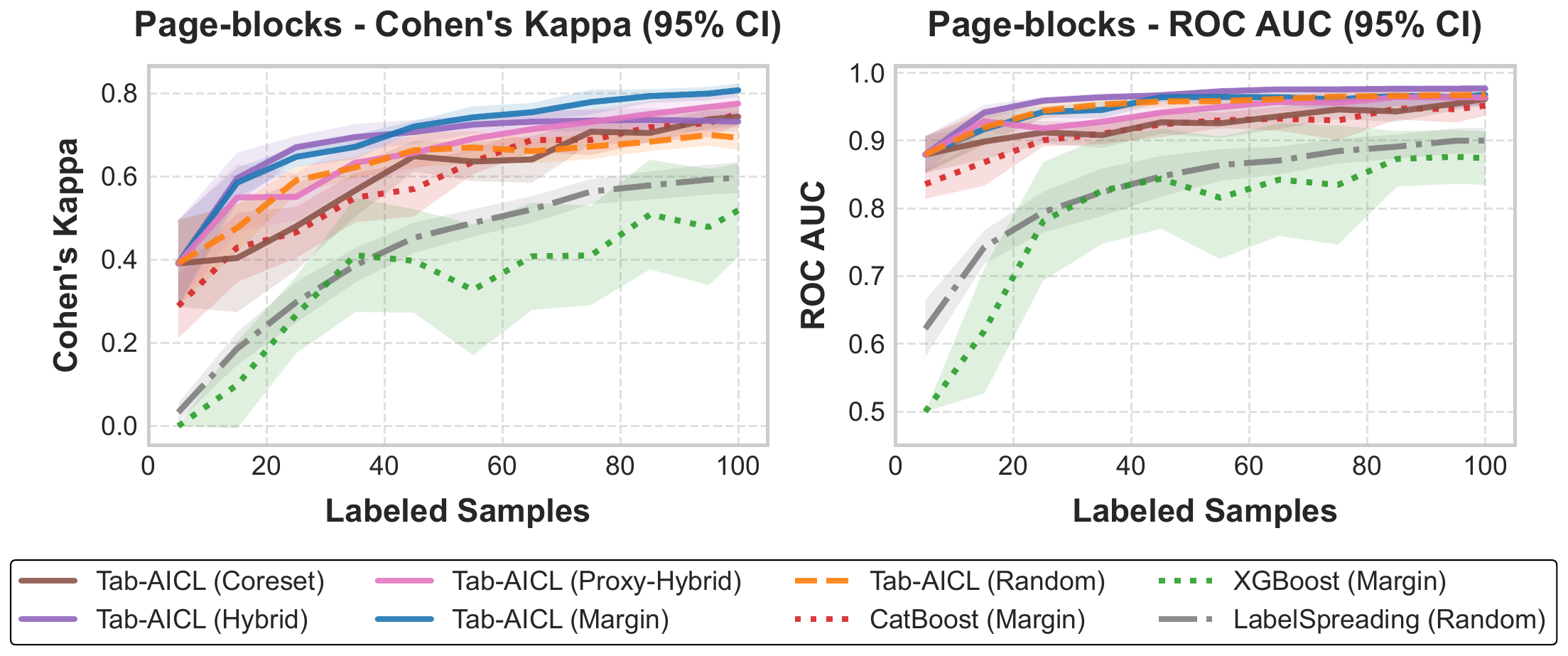}
\caption{Learning Curve for Page Blocks: The \textbf{TabPFN-Margin} strategy (blue) achieves rapid convergence, suggesting that for certain data manifolds, pure uncertainty sampling remains an effective approach.}
\label{fig:pageblocks}
\end{figure}

Figures~\ref{fig:aulc_comprehensive}--\ref{fig:pageblocks} complement the table results by showing how methods behave across the acquisition horizon.
Fig.~\ref{fig:aulc_comprehensive} summarizes the distribution of normalized AULC across the 20 datasets: hybrid rules tend to be more consistent across datasets, while pure uncertainty sampling can be more variable depending on the pool.

Inspection of representative learning curves highlights recurring patterns.
On structured datasets such as \texttt{Ionosphere} (Fig.~\ref{fig:ionosphere}), TabPFN-based methods typically reach high performance with fewer queried labels than retrained GBDT baselines in the $N<50$ regime.
On noisier benchmarks such as \texttt{Adult} (Fig.~\ref{fig:adult}), \textbf{TabPFN-Proxy-Hybrid} often exhibits tighter across-seed variability than uncertainty-only selection, consistent with the proxy stage reducing sensitivity to atypical points.
On datasets with clustered or geometric structure, \texttt{TabPFN-Hybrid} frequently maintains strong performance throughout the curve (e.g., \texttt{Vehicle}, Fig.~\ref{fig:vehicle}), suggesting that combining an uncertainty filter with a simple diversity step can be beneficial.
Finally, some acquisition rules show early differences that narrow as the labeled set grows, indicating that advantages are concentrated in the cold-start region rather than at the final budget.
Overall, the figures suggest that differences between acquisition rules are driven not only by final performance but by how quickly and how stably they improve during the $0$--$100$ label regime.

\subsection{Computational efficiency analysis}
TabPFN-based acquisition can be expensive because scoring candidates requires repeated in-context inference, whose attention cost grows quadratically with the context size. The proposed \textbf{TabPFN-Proxy-Hybrid} reduces the dominant cost by restricting TabPFN scoring to a proxy-defined shortlist. Concretely, a logistic-regression proxy computes uncertainty over the full pool and we retain only an $\alpha=0.05$ fraction (clamped as described in Section~\ref{alg:proxy}); TabPFN then evaluates uncertainty and performs diversity selection only within this subset. Relative to scoring all of $\mathcal{U}$ with TabPFN, this reduces the number of TabPFN evaluations by approximately a factor $1/\alpha$ (about $20\times$ when the clamp is inactive).

Batch size influences runtime through two opposing effects: smaller batches increase the number of acquisition iterations, while larger batches increase the context size and the cost of each TabPFN call. In our implementation, intermediate values (e.g., $B=10$) provide a practical compromise.%, and we therefore report results across multiple batch sizes rather than optimizing for a single setting.

\subsection{Ablation study: batch size}
We study the effect of the query batch size $B$ on cold-start efficiency by running the same AL protocol with $B \in \{5,10,15,20\}$ (Table~\ref{tab:batchsize}). As expected, smaller batches can improve normalized AULC because the context is updated more frequently, allowing the model to revise its predictions after fewer newly labeled points. At the same time, performance is reasonably stable across batch sizes on most datasets, with only modest degradation as $B$ increases. This suggests that Tab-AICL can be used with larger batches -- often preferred in practice to amortize annotation overhead -- without materially changing the main conclusions.

\begin{table}[b]
\caption{Batch Size Ablation (AULC $\pm$ std, TabPFN-Proxy-Hybrid)}
\label{tab:batchsize}
\resizebox{\columnwidth}{!}{%
\begin{tabular}{lcccc}
\toprule
\textbf{Dataset} & \textbf{B=5} & \textbf{B=10} & \textbf{B=15} & \textbf{B=20} \\
\midrule
Iris & 0.920 $\pm$ 0.036 & \textbf{0.931 $\pm$ 0.019} & 0.910 $\pm$ 0.016 & 0.907 $\pm$ 0.022 \\
Glass & 0.446 $\pm$ 0.059 & 0.441 $\pm$ 0.061 & \textbf{0.467 $\pm$ 0.040} & 0.442 $\pm$ 0.027 \\
Ionosphere & \textbf{0.809 $\pm$ 0.044} & 0.800 $\pm$ 0.032 & 0.784 $\pm$ 0.049 & 0.766 $\pm$ 0.092 \\
Balance-scale & \textbf{0.836 $\pm$ 0.038} & 0.811 $\pm$ 0.040 & 0.791 $\pm$ 0.037 & 0.772 $\pm$ 0.044 \\
Vehicle & \textbf{0.837 $\pm$ 0.040} & 0.811 $\pm$ 0.054 & 0.817 $\pm$ 0.040 & 0.776 $\pm$ 0.054 \\
Page-blocks & 0.651 $\pm$ 0.075 & \textbf{0.657 $\pm$ 0.043} & 0.623 $\pm$ 0.069 & 0.632 $\pm$ 0.059 \\
Parkinsons & 0.620 $\pm$ 0.114 & 0.618 $\pm$ 0.128 & 0.621 $\pm$ 0.113 & \textbf{0.625 $\pm$ 0.144} \\
Seeds & \textbf{0.911 $\pm$ 0.023} & 0.903 $\pm$ 0.022 & 0.900 $\pm$ 0.023 & 0.901 $\pm$ 0.024 \\
Bank-Marketing & 0.209 $\pm$ 0.056 & 0.217 $\pm$ 0.100 & \textbf{0.220 $\pm$ 0.086} & 0.188 $\pm$ 0.082 \\
Adult & 0.356 $\pm$ 0.076 & \textbf{0.358 $\pm$ 0.067} & 0.348 $\pm$ 0.061 & 0.340 $\pm$ 0.061 \\
CoverType & \textbf{0.341 $\pm$ 0.032} & 0.315 $\pm$ 0.046 & 0.329 $\pm$ 0.045 & 0.324 $\pm$ 0.049 \\
KC1 & 0.393 $\pm$ 0.098 & \textbf{0.412 $\pm$ 0.103} & 0.377 $\pm$ 0.122 & 0.326 $\pm$ 0.114 \\
JM1 & 0.085 $\pm$ 0.039 & 0.106 $\pm$ 0.056 & 0.097 $\pm$ 0.064 & \textbf{0.112 $\pm$ 0.041} \\
Blood-Transfusion & 0.107 $\pm$ 0.059 & \textbf{0.126 $\pm$ 0.093} & 0.109 $\pm$ 0.086 & 0.107 $\pm$ 0.087 \\
Diabetes & 0.387 $\pm$ 0.064 & \textbf{0.401 $\pm$ 0.069} & 0.368 $\pm$ 0.081 & 0.366 $\pm$ 0.071 \\
Tic-Tac-Toe & 0.158 $\pm$ 0.083 & 0.175 $\pm$ 0.075 & \textbf{0.175 $\pm$ 0.080} & 0.158 $\pm$ 0.088 \\
Credit-g & 0.132 $\pm$ 0.057 & \textbf{0.164 $\pm$ 0.058} & 0.102 $\pm$ 0.078 & 0.145 $\pm$ 0.049 \\
Steel-Plates & 0.371 $\pm$ 0.067 & 0.350 $\pm$ 0.079 & 0.373 $\pm$ 0.061 & \textbf{0.379 $\pm$ 0.046} \\
Phoneme & \textbf{0.449 $\pm$ 0.076} & 0.440 $\pm$ 0.122 & 0.390 $\pm$ 0.160 & 0.417 $\pm$ 0.147 \\
Ilpd & 0.107 $\pm$ 0.091 & 0.109 $\pm$ 0.087 & \textbf{0.131 $\pm$ 0.070} & 0.104 $\pm$ 0.080 \\
\bottomrule
\end{tabular}%
}
\vspace{0.1cm}
\parbox{\columnwidth}{\footnotesize \textbf{Remark:} Experiments conducted with fixed Proxy Filter $\alpha=0.05$.
%Analysis represents one-factor-at-a-time (OFAT) sensitivity, not a full grid search.
}
\end{table}

\begin{table*}[t]
\caption{Final Kappa Performance Summary (Mean $\pm$ std at $N=100$, Batch Size 10)}
\label{tab:kappa_performance}
\centering
\footnotesize
\resizebox{\linewidth}{!}{%
\begin{tabular}{@{}lcccccccc@{}}
\toprule
\multirow{2}{*}{\textbf{Dataset}} & \multicolumn{5}{c}{\textbf{Tab-AICL Strategies}} & \multicolumn{3}{c}{\textbf{Baselines}} \\
\cmidrule(lr){2-5} \cmidrule(lr){6-9}
 & \textbf{TabPFN-Coreset} & \textbf{TabPFN-Hybrid} & \textbf{TabPFN-Proxy-Hybrid} & \textbf{TabPFN-Margin} & \textbf{TabPFN-Random} & \textbf{CatBoost-Margin} & \textbf{XGBoost-Margin} & \textbf{LabelSpreading-Random} \\
\midrule
Iris & 0.950 $\pm$ 0.043 & 0.933 $\pm$ 0.049 & \textbf{0.950 $\pm$ 0.037} & 0.943 $\pm$ 0.047 & 0.947 $\pm$ 0.034 & 0.927 $\pm$ 0.047 & 0.880 $\pm$ 0.092 & 0.877 $\pm$ 0.060 \\
Glass & 0.564 $\pm$ 0.053 & 0.580 $\pm$ 0.060 & 0.585 $\pm$ 0.061 & \textbf{0.637 $\pm$ 0.055} & 0.592 $\pm$ 0.094 & 0.588 $\pm$ 0.091 & 0.496 $\pm$ 0.024 & 0.441 $\pm$ 0.066 \\
Ionosphere & 0.805 $\pm$ 0.078 & 0.875 $\pm$ 0.039 & \textbf{0.888 $\pm$ 0.044} & 0.878 $\pm$ 0.034 & 0.860 $\pm$ 0.037 & 0.821 $\pm$ 0.047 & 0.800 $\pm$ 0.065 & 0.574 $\pm$ 0.069 \\
Balance Scale & 0.871 $\pm$ 0.033 & 0.918 $\pm$ 0.029 & 0.940 $\pm$ 0.024 & \textbf{0.949 $\pm$ 0.018} & 0.885 $\pm$ 0.034 & 0.742 $\pm$ 0.050 & 0.756 $\pm$ 0.065 & 0.762 $\pm$ 0.035 \\
Vehicle & 0.916 $\pm$ 0.070 & 0.966 $\pm$ 0.022 & 0.980 $\pm$ 0.014 & \textbf{0.989 $\pm$ 0.009} & 0.921 $\pm$ 0.065 & 0.929 $\pm$ 0.018 & 0.897 $\pm$ 0.024 & 0.810 $\pm$ 0.053 \\
Page Blocks & 0.744 $\pm$ 0.039 & 0.732 $\pm$ 0.039 & 0.775 $\pm$ 0.056 & \textbf{0.807 $\pm$ 0.026} & 0.693 $\pm$ 0.048 & 0.751 $\pm$ 0.042 & 0.519 $\pm$ 0.181 & 0.597 $\pm$ 0.059 \\
Parkinsons & 0.692 $\pm$ 0.099 & 0.731 $\pm$ 0.141 & 0.767 $\pm$ 0.115 & \textbf{0.772 $\pm$ 0.102} & 0.677 $\pm$ 0.161 & 0.716 $\pm$ 0.129 & 0.703 $\pm$ 0.134 & 0.626 $\pm$ 0.102 \\
Seeds & 0.921 $\pm$ 0.037 & \textbf{0.936 $\pm$ 0.024} & 0.931 $\pm$ 0.031 & 0.926 $\pm$ 0.029 & 0.921 $\pm$ 0.035 & 0.879 $\pm$ 0.052 & 0.852 $\pm$ 0.040 & 0.869 $\pm$ 0.034 \\
Bank-Marketing & 0.308 $\pm$ 0.116 & 0.235 $\pm$ 0.105 & 0.280 $\pm$ 0.114 & 0.216 $\pm$ 0.079 & \textbf{0.322 $\pm$ 0.062} & 0.208 $\pm$ 0.055 & 0.150 $\pm$ 0.112 & 0.094 $\pm$ 0.032 \\
Adult & 0.388 $\pm$ 0.083 & 0.377 $\pm$ 0.072 & \textbf{0.455 $\pm$ 0.056} & 0.378 $\pm$ 0.090 & 0.408 $\pm$ 0.074 & 0.408 $\pm$ 0.069 & 0.297 $\pm$ 0.160 & 0.205 $\pm$ 0.050 \\
Covertype & 0.439 $\pm$ 0.027 & 0.445 $\pm$ 0.024 & 0.381 $\pm$ 0.049 & \textbf{0.471 $\pm$ 0.018} & 0.447 $\pm$ 0.029 & 0.420 $\pm$ 0.039 & 0.388 $\pm$ 0.038 & 0.240 $\pm$ 0.028 \\
KC1 & 0.435 $\pm$ 0.113 & 0.482 $\pm$ 0.120 & 0.455 $\pm$ 0.100 & 0.447 $\pm$ 0.132 & 0.423 $\pm$ 0.234 & \textbf{0.508 $\pm$ 0.132} & 0.292 $\pm$ 0.188 & 0.389 $\pm$ 0.116 \\
JM1 & 0.075 $\pm$ 0.050 & \textbf{0.151 $\pm$ 0.046} & 0.119 $\pm$ 0.059 & 0.061 $\pm$ 0.070 & 0.076 $\pm$ 0.056 & 0.080 $\pm$ 0.060 & 0.044 $\pm$ 0.054 & 0.097 $\pm$ 0.056 \\
Blood-Transfusion & 0.127 $\pm$ 0.107 & 0.161 $\pm$ 0.082 & 0.205 $\pm$ 0.103 & 0.197 $\pm$ 0.100 & 0.114 $\pm$ 0.086 & \textbf{0.212 $\pm$ 0.114} & 0.151 $\pm$ 0.122 & 0.143 $\pm$ 0.063 \\
Diabetes & 0.434 $\pm$ 0.042 & 0.447 $\pm$ 0.062 & \textbf{0.454 $\pm$ 0.058} & 0.417 $\pm$ 0.048 & 0.429 $\pm$ 0.051 & 0.372 $\pm$ 0.061 & 0.290 $\pm$ 0.149 & 0.242 $\pm$ 0.090 \\
Tic-Tac-Toe & 0.090 $\pm$ 0.098 & 0.189 $\pm$ 0.056 & 0.203 $\pm$ 0.065 & 0.181 $\pm$ 0.098 & \textbf{0.233 $\pm$ 0.109} & 0.180 $\pm$ 0.079 & 0.163 $\pm$ 0.107 & 0.111 $\pm$ 0.049 \\
Credit-g & 0.229 $\pm$ 0.096 & \textbf{0.261 $\pm$ 0.073} & 0.251 $\pm$ 0.070 & 0.236 $\pm$ 0.079 & 0.209 $\pm$ 0.104 & 0.225 $\pm$ 0.071 & 0.122 $\pm$ 0.113 & 0.156 $\pm$ 0.049 \\
Steel-Plates & 0.451 $\pm$ 0.068 & \textbf{0.617 $\pm$ 0.019} & 0.411 $\pm$ 0.108 & 0.592 $\pm$ 0.030 & 0.574 $\pm$ 0.034 & 0.510 $\pm$ 0.037 & 0.465 $\pm$ 0.042 & 0.468 $\pm$ 0.040 \\
Phoneme & 0.517 $\pm$ 0.055 & 0.563 $\pm$ 0.040 & 0.555 $\pm$ 0.091 & \textbf{0.607 $\pm$ 0.028} & 0.588 $\pm$ 0.029 & 0.578 $\pm$ 0.037 & 0.456 $\pm$ 0.096 & 0.496 $\pm$ 0.051 \\
Ilpd & 0.023 $\pm$ 0.041 & 0.086 $\pm$ 0.099 & 0.133 $\pm$ 0.106 & 0.096 $\pm$ 0.058 & \textbf{0.185 $\pm$ 0.104} & 0.156 $\pm$ 0.059 & 0.093 $\pm$ 0.095 & 0.134 $\pm$ 0.075 \\
\bottomrule
\end{tabular}%
}
\vspace{0.2cm}
\parbox{\linewidth}{
\footnotesize \textbf{Note:} Bold indicates the strategy with the highest mean Kappa per dataset. All values represent Mean $\pm$ Standard Deviation across 10 random seeds.
}
\end{table*}

\begin{table*}[t]
\caption{Final ROC AUC Performance Summary (Mean $\pm$ std at $N=100$, Batch Size 10)}
\label{tab:auc_performance}
\centering
\footnotesize
\resizebox{\linewidth}{!}{%
\begin{tabular}{@{}lcccccccc@{}}
\toprule
\multirow{2}{*}{\textbf{Dataset}} & \multicolumn{5}{c}{\textbf{Tab-AICL Strategies}} & \multicolumn{3}{c}{\textbf{Baselines}} \\
\cmidrule(lr){2-5} \cmidrule(lr){6-9}
 & \textbf{TabPFN-Coreset} & \textbf{TabPFN-Hybrid} & \textbf{TabPFN-Proxy-Hybrid} & \textbf{TabPFN-Margin} & \textbf{TabPFN-Random} & \textbf{CatBoost-Margin} & \textbf{XGBoost-Margin} & \textbf{LabelSpreading-Random} \\
\midrule
Iris & \textbf{0.997 $\pm$ 0.003} & 0.995 $\pm$ 0.005 & 0.996 $\pm$ 0.003 & 0.996 $\pm$ 0.004 & 0.996 $\pm$ 0.004 & 0.994 $\pm$ 0.006 & 0.956 $\pm$ 0.044 & 0.986 $\pm$ 0.009 \\
Glass & 0.921 $\pm$ 0.026 & 0.929 $\pm$ 0.018 & 0.926 $\pm$ 0.016 & \textbf{0.936 $\pm$ 0.013} & 0.929 $\pm$ 0.033 & 0.907 $\pm$ 0.022 & 0.849 $\pm$ 0.017 & 0.854 $\pm$ 0.041 \\
Ionosphere & 0.973 $\pm$ 0.009 & 0.984 $\pm$ 0.007 & 0.985 $\pm$ 0.006 & \textbf{0.987 $\pm$ 0.006} & 0.981 $\pm$ 0.008 & 0.964 $\pm$ 0.016 & 0.947 $\pm$ 0.018 & 0.899 $\pm$ 0.044 \\
Balance Scale & 0.989 $\pm$ 0.003 & 0.994 $\pm$ 0.003 & 0.996 $\pm$ 0.002 & \textbf{0.998 $\pm$ 0.002} & 0.990 $\pm$ 0.006 & 0.956 $\pm$ 0.012 & 0.955 $\pm$ 0.030 & 0.957 $\pm$ 0.011 \\
Vehicle & 0.996 $\pm$ 0.003 & 0.999 $\pm$ 0.001 & 1.000 $\pm$ 0.000 & \textbf{1.000 $\pm$ 0.000} & 0.998 $\pm$ 0.001 & 0.995 $\pm$ 0.003 & 0.986 $\pm$ 0.008 & 0.975 $\pm$ 0.011 \\
Page Blocks & 0.962 $\pm$ 0.011 & \textbf{0.977 $\pm$ 0.006} & 0.964 $\pm$ 0.022 & 0.968 $\pm$ 0.025 & 0.968 $\pm$ 0.012 & 0.952 $\pm$ 0.024 & 0.874 $\pm$ 0.064 & 0.900 $\pm$ 0.030 \\
Parkinsons & 0.932 $\pm$ 0.055 & \textbf{0.958 $\pm$ 0.040} & 0.940 $\pm$ 0.077 & 0.956 $\pm$ 0.035 & 0.919 $\pm$ 0.053 & 0.935 $\pm$ 0.036 & 0.925 $\pm$ 0.047 & 0.916 $\pm$ 0.036 \\
Seeds & 0.996 $\pm$ 0.003 & 0.995 $\pm$ 0.003 & \textbf{0.996 $\pm$ 0.003} & 0.996 $\pm$ 0.003 & 0.995 $\pm$ 0.004 & 0.982 $\pm$ 0.009 & 0.975 $\pm$ 0.014 & 0.986 $\pm$ 0.005 \\
Bank-Marketing & 0.836 $\pm$ 0.033 & 0.837 $\pm$ 0.025 & 0.830 $\pm$ 0.036 & 0.799 $\pm$ 0.057 & \textbf{0.854 $\pm$ 0.011} & 0.759 $\pm$ 0.047 & 0.705 $\pm$ 0.062 & 0.655 $\pm$ 0.031 \\
Adult & 0.841 $\pm$ 0.024 & 0.820 $\pm$ 0.042 & 0.833 $\pm$ 0.030 & 0.830 $\pm$ 0.051 & \textbf{0.851 $\pm$ 0.027} & 0.817 $\pm$ 0.023 & 0.778 $\pm$ 0.056 & 0.658 $\pm$ 0.033 \\
Covertype & \textbf{0.904 $\pm$ 0.011} & 0.898 $\pm$ 0.008 & 0.886 $\pm$ 0.014 & 0.902 $\pm$ 0.008 & 0.899 $\pm$ 0.009 & 0.841 $\pm$ 0.010 & 0.786 $\pm$ 0.017 & 0.761 $\pm$ 0.028 \\
KC1 & 0.817 $\pm$ 0.071 & 0.812 $\pm$ 0.041 & 0.824 $\pm$ 0.055 & 0.816 $\pm$ 0.039 & 0.814 $\pm$ 0.094 & \textbf{0.827 $\pm$ 0.046} & 0.798 $\pm$ 0.046 & 0.801 $\pm$ 0.044 \\
JM1 & 0.643 $\pm$ 0.076 & \textbf{0.673 $\pm$ 0.027} & 0.645 $\pm$ 0.034 & 0.638 $\pm$ 0.075 & 0.670 $\pm$ 0.035 & 0.626 $\pm$ 0.041 & 0.621 $\pm$ 0.057 & 0.537 $\pm$ 0.039 \\
Blood-Transfusion & 0.717 $\pm$ 0.058 & \textbf{0.721 $\pm$ 0.045} & 0.720 $\pm$ 0.030 & 0.705 $\pm$ 0.047 & 0.697 $\pm$ 0.046 & 0.656 $\pm$ 0.049 & 0.660 $\pm$ 0.044 & 0.609 $\pm$ 0.027 \\
Diabetes & 0.819 $\pm$ 0.018 & \textbf{0.827 $\pm$ 0.021} & 0.820 $\pm$ 0.017 & 0.814 $\pm$ 0.020 & 0.819 $\pm$ 0.020 & 0.761 $\pm$ 0.042 & 0.730 $\pm$ 0.048 & 0.723 $\pm$ 0.036 \\
Tic-Tac-Toe & 0.680 $\pm$ 0.027 & 0.681 $\pm$ 0.038 & 0.668 $\pm$ 0.033 & 0.651 $\pm$ 0.068 & \textbf{0.722 $\pm$ 0.023} & 0.648 $\pm$ 0.042 & 0.638 $\pm$ 0.045 & 0.602 $\pm$ 0.029 \\
Credit-g & 0.712 $\pm$ 0.030 & \textbf{0.752 $\pm$ 0.027} & 0.746 $\pm$ 0.021 & 0.724 $\pm$ 0.024 & 0.743 $\pm$ 0.031 & 0.679 $\pm$ 0.051 & 0.645 $\pm$ 0.043 & 0.648 $\pm$ 0.040 \\
Steel-Plates & 0.889 $\pm$ 0.019 & 0.906 $\pm$ 0.017 & 0.891 $\pm$ 0.015 & \textbf{0.906 $\pm$ 0.024} & 0.900 $\pm$ 0.024 & 0.884 $\pm$ 0.015 & 0.846 $\pm$ 0.029 & 0.855 $\pm$ 0.024 \\
Phoneme & 0.829 $\pm$ 0.024 & 0.849 $\pm$ 0.020 & 0.826 $\pm$ 0.030 & 0.864 $\pm$ 0.014 & \textbf{0.865 $\pm$ 0.013} & 0.842 $\pm$ 0.018 & 0.816 $\pm$ 0.026 & 0.810 $\pm$ 0.019 \\
Ilpd & 0.727 $\pm$ 0.025 & 0.718 $\pm$ 0.032 & 0.704 $\pm$ 0.037 & 0.676 $\pm$ 0.052 & \textbf{0.729 $\pm$ 0.021} & 0.661 $\pm$ 0.054 & 0.636 $\pm$ 0.061 & 0.633 $\pm$ 0.026 \\
\bottomrule
\end{tabular}%
}
\vspace{0.2cm}
\parbox{\linewidth}{
\footnotesize \textbf{Note:} Bold indicates the strategy with the highest mean ROC AUC per dataset. All values represent Mean $\pm$ Standard Deviation across 10 random seeds.
}
\end{table*}

\section{Limitations, trade-offs, and conclusion}

Tab-AICL reframes active learning for tabular foundation models as \emph{context optimization}: at each step, we update the labeled context provided to TabPFN rather than retraining task-specific parameters. Across 20 classification benchmarks (10 seeds), TabPFN-based methods are strong cold-start predictors, and in many cases active acquisition improves normalized AULC up to 100 labels relative to both random querying and retrained gradient-boosting baselines under the same protocol. However, the results also show clear trade-offs and regime dependence.

\paragraph{Limitations and trade-offs}
First, Tab-AICL is inherently a \emph{small-data} approach: inference scales quadratically with the context length and is bounded by the model maximum context window, so benefits can saturate as the labeled set grows and costs increase. Second, the very early phase remains sensitive to initialization: with only a few labeled points, class imbalance or unlucky seeds can bias the trajectory before acquisition rules have enough signal to correct it. Third, acquisition rules can fail in predictable ways. Uncertainty-based selection can over-focus on atypical or noisy points; diversity-only selection can waste labels away from informative regions; and proxy filtering introduces an additional modeling assumption that may be lossy on complex tasks. %Empirically, this is reflected by datasets where TabPFN-Random is competitive with, or better than, active querying in the 0--100 label regime.

\paragraph{Practical implication}
These constraints suggest using Tab-AICL as a \emph{cold-start module}: exploit TabPFN (optionally with active acquisition) to rapidly reach a competent model with tens to a few hundred labels, then hand off to scalable tabular learners (e.g., CatBoost/XGBoost) when additional labels are available and retraining costs become acceptable.

\paragraph{Future directions}
Two concrete extensions follow naturally: (i) automatic criteria that decide when to transition from context-based inference to retrained models, and (ii) context compression/distillation methods that transfer the information in the acquired context into a lightweight student, mitigating the quadratic inference cost. Meta-selection of acquisition rules could also be designed, using early iterations to choose between uncertainty-, diversity-, and proxy-based strategies rather than committing to a single rule a priori.
While this work focuses on classification, future research should extend Tab-AICL to regression and multi-label tabular problems, adapting the acquisition functions and potentially developing specialized foundation models for these tasks.

\section*{Acknowledgments}
This paper is supported by PNRR-PE-AI FAIR project funded by the NextGeneration EU program.

\appendices

\section{Final-step metrics at the 100-label budget}
\label{app:final_metrics}
We report final-step performance at the budget limit ($|\mathcal{L}|=100$) for all methods, using Cohen's $\kappa$ (Table~\ref{tab:kappa_performance}) and ROC AUC (one-vs-rest macro-average for multi-class; Table~\ref{tab:auc_performance}), complementing the normalized AULC results in the main text.

\bibliography{thebibliography}
\bibliographystyle{IEEEtran}

\end{document}